\documentclass{bmvc2k}

\title{Advancing Medical Image Segmentation:\\
Morphology-Driven Learning with Diffusion
Transformer}

\addauthor{Sungmin Kang}{rkdtjdals97@dgu.ac.kr}{1}
\addauthor{Jaeha Song}{archiiive99@gmail.com}{2}
\addauthor{Jihie Kim}{jihie.kim@dgu.edu}{1}

\addinstitution{
 Department of Computer Science and Artificial Intelligence,\\
 Dongguk University,\\
 Seoul, Korea
}
\addinstitution{
 Department of Computer Science and Engineering,\\
 Dongguk University,\\
 Seoul, Korea
}

\runninghead{KANG ET AL.}{Morphology-Driven Learning with Diffusion Transformer}

\usepackage{import}
\usepackage{times}
\usepackage{soul}
\usepackage{dsfont}
\usepackage{url}
\usepackage{multirow} 
\usepackage{wrapfig}
\usepackage{graphicx}
\usepackage{amsmath}
\usepackage{amsthm}
\usepackage{amsfonts}
\usepackage{amssymb}
\usepackage{booktabs}
\usepackage{algorithm}
\usepackage{algorithmic}
\usepackage[switch]{lineno}
\usepackage{xcolor}
\begin{document}

\maketitle

\begin{abstract}
Understanding the morphological structure of medical images and precisely segmenting the region of interest or abnormality is an important task that can assist in diagnosis. However, the unique properties of medical imaging make clear segmentation difficult, and the high cost and time-consuming task of labeling leads to a coarse-grained representation of ground truth. Facing with these problems, we propose a novel \textbf{Diffusion Transformer Segmentation (DTS)} model for robust segmentation in the presence of noise.
We propose an alternative to the dominant Denoising U-Net encoder through experiments applying a transformer architecture, which captures global dependency through self-attention. Additionally, we propose k-neighbor label smoothing, reverse boundary attention, and self-supervised learning with morphology-driven learning to improve the ability to identify complex structures. Our model, which analyzes the morphological representation of images, shows better results than the previous models in various medical imaging modalities, including CT, MRI, and lesion images. Our code and dataset are publicly available at: \url{https://github.com/ready2drop/DTS}
\end{abstract}

\section{Introduction}
Medical image segmentation is crucial in improving our understanding of complex anatomy, providing critical insights for accurate medical diagnosis and precise treatment planning. This is especially important in computed tomography(CT) scans, where the intrinsic complexity of medical images presents unique challenges that require sophisticated solutions for organ segmentation. Unlike general images, CT images are quantitative imaging, and pixel intensities are normalized to Hounsfield units (HU) values\cite{LEV2002427}. (\textit{e.g.,} air as \textit{-1000} HU, bone as \textit{+400 to +1000} HU). Therefore, clinicians must understand the quantitative meanings and select the appropriate range to enhance the visual contrast of specific tissues or organs. In particular, research is conducted to find appropriate range values for each tissue or organ in CT scans\cite{LEV2002427, raybaud2020principles, ADAMS2012277, SAHI201471, mertens2023use} and studies show that segmenting CT images with an inappropriate HU range normalized leads to poor performance\cite{Huo_2019,kim2020new}. \vspace{-0.4cm}
\begin{wrapfigure}{r}{5.8cm}
\begin{tabular}{cc}
\centering
\bmvaHangBox{{\includegraphics[width=2.5cm]{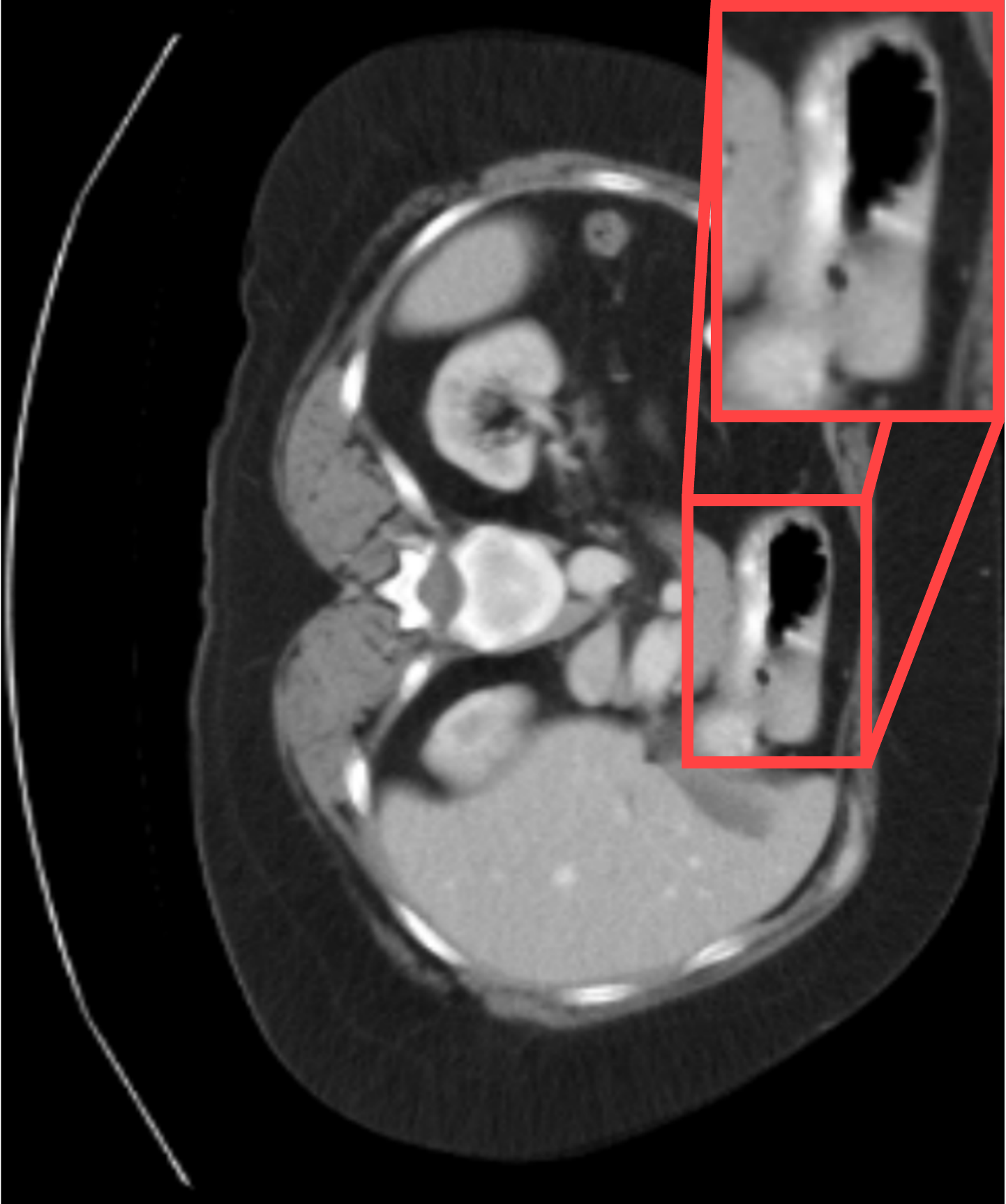}}}&
\bmvaHangBox{{\includegraphics[width=2.5cm]{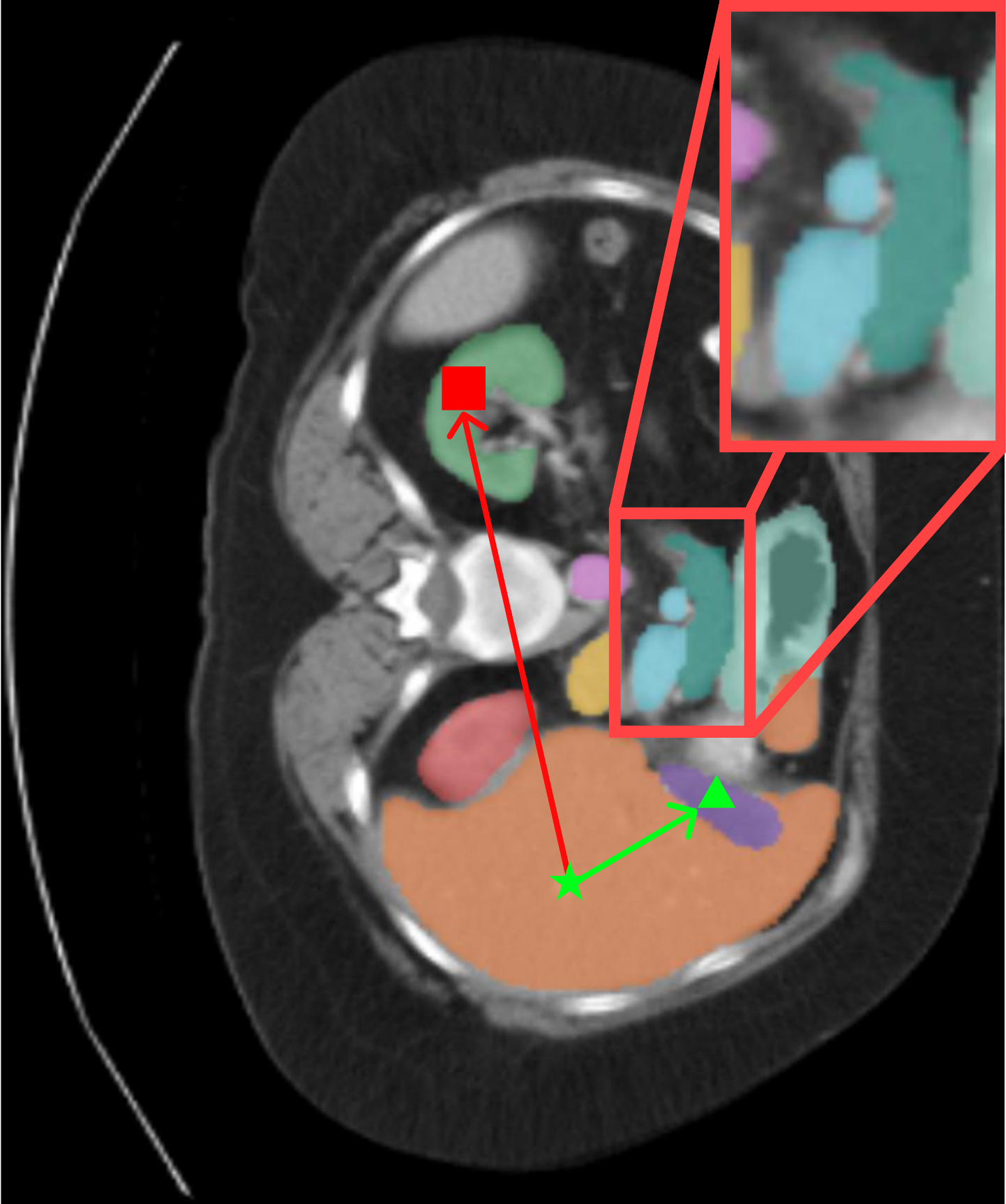}}}\\
(a)&(b)
\end{tabular}
\vspace{-0.3cm}
\caption{\color{violet}Medical imaging drawbacks.}
\label{fig:fig1}
\vspace{-0.4cm}
\end{wrapfigure}

 This is because inappropriate normalization can occlude organs, as illustrated in Fig.\ref{fig:fig1} (a). Additionally, clinicians can have different opinions in labeling\cite{alpert2004quality, monteiro2020stochastic, becker2019variability, joskowicz2019inter}. Due to this, ground truths are not determinstic, and it may be difficult to obtain detailed representations of organ or lesion labels. Inaccurate manual labeling can further increase the complexity of organ segmentation, as shown in Fig.\ref{fig:fig1} (b). We address intrinsic challenges in medical images with an architecture that combines the advantages of the adaptive and resilient Swin Transformer encoder\cite{liu2021swin} with the efficient decoder in UNet\cite{ronneberger2015unet}. We break away from the conventional Denoising U-Net\cite{rombach2022highresolution} structure because we need a model that captures a global contextual representation and can handle the various medical imaging data. In addition, we introduce three approaches to improve the segmentation process further. First, distance-aware label smoothing\cite{yu2017transfer, zhang2023soft, Felfeliyan_2023} is a guidance mechanism that recognizes anatomical locations in the medical image and smoothes labels by calculating location-based distances. Second, reverse boundary attention captures areas of subtle and ambiguous boundaries. This component contributes to more precise and accurate segmentation by explicitly directing the model attention to edges\cite{wang2022boundary, lee2020structure}, especially in the regions that have not been manually labeled. Third, self-supervised learning\cite{alzubaidi2020towards, DBLP:journals/corr/abs-2201-01283, shurrab2022self} allows complex features of organs to capture meaningful representations from input images in a scenario of insufficient data. We reduce reliance on labeled data and improve model adaptability to diverse and complex features of medical images. Our proposed method demonstrates generalizability beyond medical images when we evaluate it with a different domain task that can utilize morphological information. Therefore, our contribution is summarized as follows.
\begin{itemize}
    \vspace{-0.2cm}
    \item  We presents a new \textbf{diffusion transformer segmentation(DTS)} model which performs better than previous framework(\textit{i.e.} CNN based Denoising Diffusion Probabilistic Model).
    \vspace{-0.3cm}
    \item We introduce a novel approach to address the medical image segmentation by integrating \textbf{morphology-driven learning} into the image processing, such as k-neighbor label smoothing, reverse boundary attention, and self-supervised learning.
    \vspace{-0.3cm}
    \item Our model demonstrates the generality in segmentation tasks in medical modalities such as CT,  MRI, and lesion images and further suggests that this approach may be adaptable to other domains.
    \vspace{-0.3cm}
\end{itemize}

\section{Related Work} \label{sec:related_work}

\noindent \textbf{The diffusion segmentation model} which applies the generative diffusion process, allows users to manipulate the ambiguity of each time step through a hierarchical structure, solving the image quality and diversity problems of existing methods, allowing the learning process to proceed stably. There is research that has notable potential applied to medical imaging\cite{kim2022diffusion, guo2023accelerating, wolleb2022diffusion}. SegDiff\cite{amit2022segdiff}, which showed consistent performance under various imaging conditions, is the first approach to solving the image segmentation problem by applying diffusion. The feature of this model is a mechanism that integrates the information of the input image and the current estimate of the segmentation map through each encoder and uses the decoder to improve the segmentation map iteratively. MedSegDiff\cite{wu2023medsegdiff} also applied the diffusion segmentation model to medical image segmentation. The input of the conditional image and noise segmentation map are integrated using a mechanism such as SegDiff, but high-frequency noise is constrained through the Fast Fourier transform module during the connection process. In addition, Diff-UNet\cite{xing2023diffunet} implemented the standard U-shaped architecture, which learns from the input volume in medical image segmentation effectively to extract semantic information. Here, we focus on the architecture and compare it with the existing diffusion segmentation model to demonstrate through experiments that the inductive bias, which is a major feature of CNN, can be replaced by ViT in diffusion segmentation.

\noindent \textbf{Label Smoothing for Image Segmentation.} Ground truth labeling for image segmentation is a time-consuming and intensive task involving experts. These processes are inherently subjective and susceptible to factors such as image quality, observer diversity, and difficulty depicting specific structures. Moreover, earlier label smoothing methods\cite{DBLP:journals/corr/SzegedyVISW15,müller2020does, 10.1167/tvst.9.2.34,islam2021spatially, DBLP:journals/corr/abs-2011-12562, liu2022devil}, the inter-class relationships are usually overlooked since the labels are smoothed into one-hot encoding vectors. To address these challenges, we experimentally highlight the implementation of strategic label smoothing based on the spatial location of organs.

\noindent \textbf{Reverse boundary attention} refers to the integration of reverse attention\cite{DBLP:journals/corr/HuangXWLWSK17, Lou_2023}, which learns opposite concepts that are not associated with the target class in a way that substitutes existing attention mechanisms for objects, and boundary attention\cite{ghadimi2015skull, Bai2016AutomaticWH, 1530294}, which emphasizes pixels or features of parts related to the boundary. This mechanism plays a crucial role in enhancing the performance of object segmentation in medical images. This is particularly important because medical imaging, such as CT and MRI scans, often exhibit ambiguous organ boundaries and significant amounts of noise, posing challenges for accurate segmentation. Therefore, we explore the benefits of combining unique advantages, such as a reverse boundary attention mechanism, into our framework.

\section{DTS: Diffusion Transformer Segmentation}
 The diffusion model is a generative model that consists of two stages: a diffusion process and a denoising process. In the diffusion process, Gaussian noise is added incrementally to the segmentation label $x_0$ over a series of steps $t$. 
 
\vspace{-0.3cm}
\begin{equation}
p_\theta(x_{0:T-1} | x_T) := \prod_{t=1}^{T} p_\theta(x_{t-1} | x_t) 
\label{eq:eq1}
\end{equation}

\vspace{-0.2cm}
\begin{equation}
  p_\theta(x_{t-1} | x_t) := \mathcal{N}(x_{t-1}; \mu_\theta(x_t, t), \Sigma_\theta(x_t, t))
\label{eq:eq2}
\end{equation}
\vspace{-0.3cm}

The denoising process, parametrized by $\theta$, involves training a neural network to recover the original data from the noise, and the distribution $p_\theta (x_t)$ is defined as $\mathcal{N}(x_T; 0, I_{n \times n})$ where $I$ represents the raw image assumed to be an $n\times n$ matrix. The denoising process then operates to transform the latent variable distribution $p_\theta (x_t)$ (\textit{i.e.} gaussian noise image) into the data distribution $p_\theta (x_0)$ (\textit{i.e.} final segmentation map).
\begin{figure}[t!]
    \centering
    \includegraphics[keepaspectratio=True, width=\linewidth]{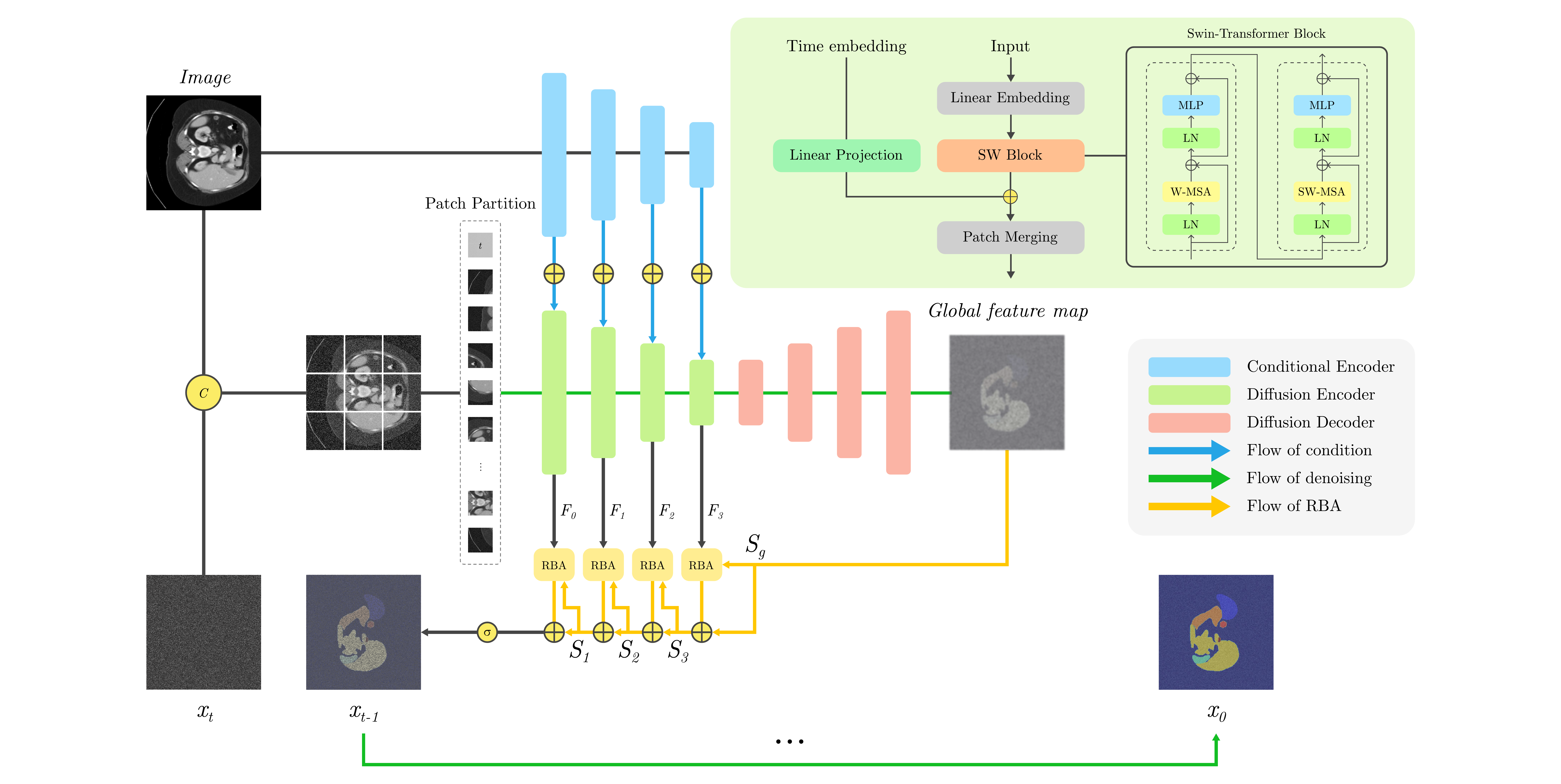}
    \vspace{-0.5cm}
    \caption{\color{violet}Overview of our proposed diffusion transformer segmentation(DTS) model.} 
    \label{fig:fig2}
    \vspace{-0.5cm}
\end{figure}

\begin{wrapfigure}{r}{6cm}
\centering
\includegraphics[width=6cm]{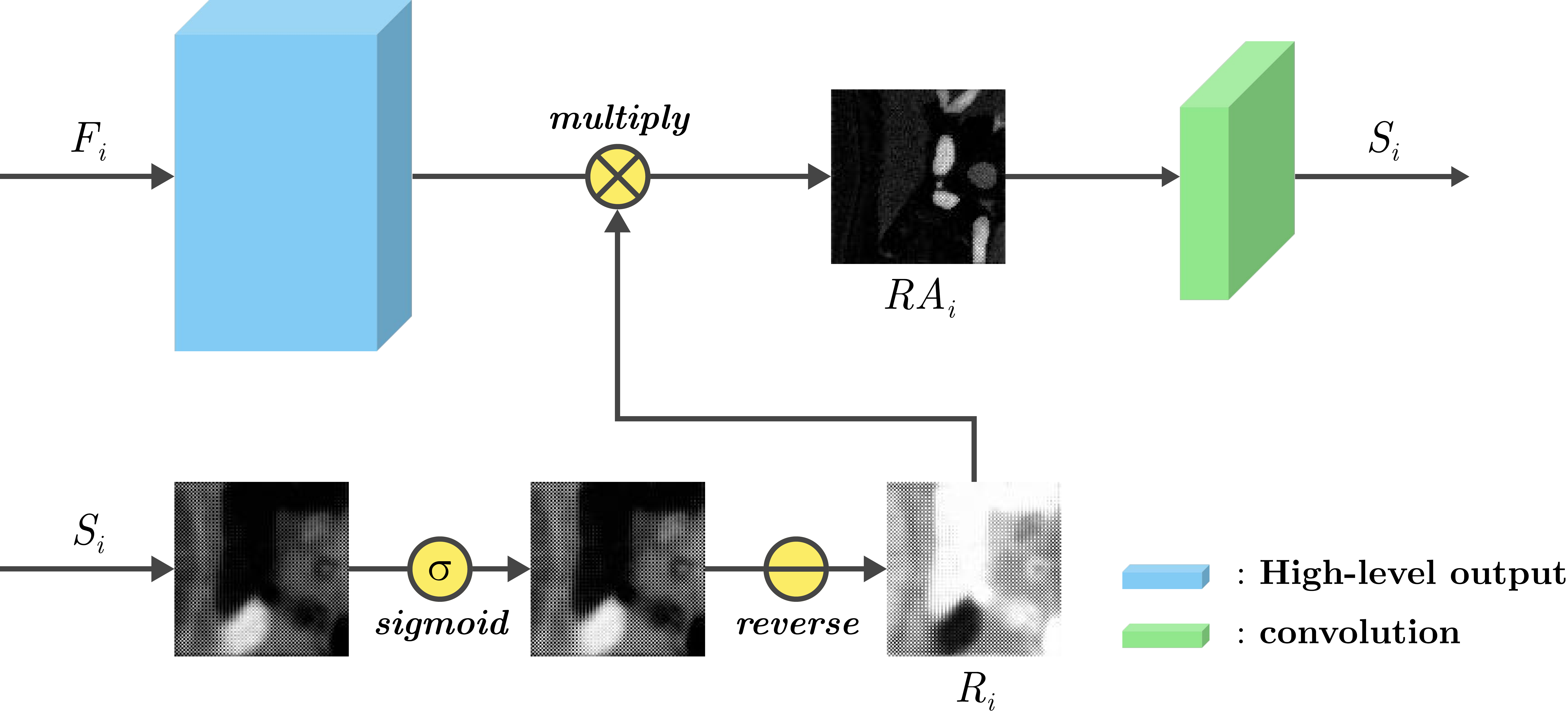}
\vspace{-0.7cm}
\caption{\color{violet}Illustration of the RBA module.}
\label{fig:fig3}
\vspace{-0.5cm}
\end{wrapfigure}

\begin{wraptable}{l}{6.5cm}
\centering
\scriptsize
\vspace{-0.3cm}
\resizebox{\linewidth}{!}{%
\begin{tabular}{cc}
\multirow{2}{*}{Architecture} & Average Accuracy \\ \cline{2-2}
                                    & Dice $\uparrow$  \\ \hline
Denoising U-Net\cite{rombach2022highresolution} & $79.74 \pm 0.30 $    \\
\textbf{Ours(DTS)*} & $\textbf{81.12}\pm 0.19 $  \\ \hline
\end{tabular}
}
\vspace{-0.6mm}
\caption{\color{violet}Comparison of an encoder network on the ISIC dataset.}
\label{table:encoder_ablation_table}
\vspace{-0.7cm}
\end{wraptable} The denoising phase shown in Fig.\ref{fig:fig2} follows the encoder-decoder network structure of standard denoising autoencoder\cite{rombach2022highresolution}. As shown in Table.\ref{table:encoder_ablation_table}, we empirically suggest the possibility of replacing the latent diffusion encoder with a Swin transformer\cite{liu2021swin}, which has advantages such as scalability and computational efficiency when processing various images due to its hierarchical structure. Also, similar to the conditional mechanism\cite{mirza2014conditional, sohn2015learning, rombach2022highresolution}, our model incorporates another type of conditional encoder, $\tau_{\theta}$ where the original image is used as input. These are demonstrated in the \textbf{\color{lime}diffusion encoder} and \textbf{\color{cyan}conditional encoder} in Fig.\ref{fig:fig2}. Our method combines information from the current estimate $x_t$, the image $I$, and the time step index $t$ to adjust the step estimate function ${\epsilon}_{\theta}$ at the input. It also takes the conditional image $\tau_{\theta}(I)$) and reconstructs it through a UNet decoder to produce the global feature map. Subsequently, the RBA modules facilitate the derivation of the final segmentation map, which exhibits precise edge representation, as detailed in Fig.\ref{fig:fig3}. In conclusion, $DTS(\cdot)$ represents our novel diffusion transformer segmentation model, which performs segmentation by integrating the described components.

\vspace{-0.2cm}
 \begin{equation}
  {\epsilon}_{\theta}(x_t,I,t) = DTS((x_t,I),t, \tau_{\theta}(I))
 \label{eq:eq5}
 \end{equation}

\section{Morphology Driven Learning}

\noindent \textbf{$k$-Neighbor Label smoothing by organ distance.}
We explore medical data from body parts such as the abdomen and brain, which have organs or diseases located structurally within a compact space. As the relative positions of organs do not differ from person to person, \vspace{-0.4cm}
\begin{wraptable}{l}{5.5cm}
\centering
\scriptsize
\resizebox{\linewidth}{!}{%
\vspace{-0.5cm}
\begin{tabular}{lccc}
\multirow{2}{*}{Label smoothing}  &  \multicolumn{3}{c}{Average Accuracy}\\
\cmidrule{2-4}
&  Dice $\uparrow$ & HD $\downarrow$ \\
\midrule
 \ \ \ \ \ Scratch & $81.12$ & $5.11$   \\
\midrule
 \ \ \ \ \ $k$-NLS   \\
 \ \ \ \ \ $\mathcal{\alpha\text{= 0.1}}$ & $\textbf{84.41}$ & $\textbf{4.17}$ \\
 \ \ \ \ \ $\mathcal{\alpha\text{= 0.2}}$ & $84.35$ & $4.20$ \\
 \ \ \ \ \ $\mathcal{\alpha\text{= 0.3}}$ & $83.31$ & $4.53$ \\
\bottomrule
\end{tabular}
}
\vspace{-0.3mm}
\caption{\color{violet}Accuracy changes with different $\alpha$(scale factor) values.}
\vspace{-0.4cm}
\label{table:ls_scalefactor_ablation_table}
\end{wraptable}

 we propose a $k$-neighbor label smoothing method that leverages the relative positions of organs for distance-aware smoothing of the labels of $k$ neighbors for a given class or organ. In a multi-class ($k>2$) situation, such as in this case, there is an advantage if there is a positional relationship between them. The positional relationship refers to the relative positional relationship of organs anatomically. As shown in Fig.~\ref{fig:fig1} (b), the \textbf{\color{orange}liver($\star$)} is close to the \textbf{\color{violet}gall bladder($\blacktriangle$)} but relatively far from the \textbf{\color{teal}left kidney($\blacksquare$)}. We provided semantic information to the model based on which body structure would match this prior knowledge.  
The equation of k-neighbor label smoothing ($k-NLS$) is:
\vspace{-0.2cm}
\begin{equation}
d_t = \{d_{xyz} \mid x, y, z \in N, x < W, y < H, z < D \} \\
 \label{eq:eq6}
\end{equation}
The distance is calculated channel-wise, measuring the distance between a random point and the centroid of $i$th class.
\vspace{0.1cm}
\begin{equation}
y_t^{k-NLS} = \left|y_t- \frac{\alpha}{d_{t} +\epsilon} \right| \\    
\label{eq:eq7}
\end{equation}

where $y_t$ is "$1$" for the target class and "$0$" for the rest of all, the label smoothing scale factor $\alpha$ is crucial. Based on previous research\cite{müller2020does} and empirical experiments on the BTCV dataset(Table.\ref{table:ls_ablation_table}), opting for $\alpha$ as 0.1 yields optimal outcomes. $\epsilon$ is constant $1e^{-6}$ to prevent division by zero, and $d_{x,y,z} = \{d_0,d_1,...,d_i \mid i=k\}$ is a set of centroids and distances between each pixel and class.
The pseudo code is expressed as follows:
\vspace{-0.2cm}
\begin{algorithm}
    \caption{K-Neighbor Label smoothing}
    \label{alg:algorithm}
    \textbf{Input}: label\\
    \textbf{Parameter}: $\epsilon$(constant factor), $\alpha$(scale factor)\\
    \textbf{Output}: smoothed label
    \begin{algorithmic}[1]
        \STATE Let encoded\_label = one\_hot\_encoding(label)
        \STATE Let coordinates = meshgrid(W, H, D)
        \STATE Let centroids = compute\_centroids(encoded\_label)
        \STATE Let d = tensor of shape [C, W, H, D]
        \FOR{each class c in range(C)}
            \FOR{each (x, y, z) in coordinates}
                \STATE d[c, x, y, z] = distance((x, y, z), centroids[:, c])
            \ENDFOR
        \ENDFOR
        \STATE Let smoothed\_label = abs(label - $\alpha$ / (d + $\epsilon$)) 
        \RETURN smoothed\_label
    \end{algorithmic}
\end{algorithm}
\vspace{-0.2cm}

\noindent \textbf{RBA: Reverse-boundary Attention.}
Complex anatomy and the inherent ambiguity in defining boundaries of adjacent organs are factors that hinder accurate segmentation of organ boundaries in medical images. Considering that these factors are likely to result in false positives or missing details in the initial segmentation, our approach includes selectively dropping or reducing the prediction weights of overlooked regions. The Reverse Boundary Attention method aims to improve the prediction of segmentation models by gradually capturing and specifying areas that may have been initially ambiguous. Thus, our architecture removes previously estimated predictive areas from high-level output features where existing estimates are upsampled in deeper layers, sequentially explores these details, including areas and boundaries, and finally, improves the segmentation model predictions progressively.
In the Fig.~\ref{fig:fig2}, the global feature map which is the output of the decoder, is resized to match the input size using a convolution layer, and reverse attention\cite{huang2017semantic} is then performed to obtain the weight $R_i$. Multiplying(element-wise $\odot$) this by the high-level output$\{F_i ,i=0,1,2,3\}$ to obtain the output reverse attention $RA_i$.
\begin{equation}
 R_i = \ominus(\sigma(\mathcal{U}(S_{i+1})))
\label{eq:eq9}
\end{equation}

\vspace{-0.5cm}
\begin{equation}
RA_i = F_i \odot R_i.
\label{eq:eq8}
\end{equation}
\vspace{-0.5cm}
 
 where $\mathcal{U}(\cdot)$, $\sigma(\cdot)$, $\ominus(\cdot)$ is up-sampling, sigmoid, reverse function respectively,
 The reverse function removes the matrix, which in all the elements is $1$. 

 As shown below, the reverse attention weight $RA_i$ is passed through two convolution layers with normalization and finally the reverse boundary attention $S_{i+1}$ is obtained.
 
 \vspace{-0.3cm}
 \begin{equation} 
 S_{i+1} = L_{conv}(RA_i)
 \label{eq:eq10}
 \end{equation}
 \vspace{-0.3cm}

\begin{wrapfigure}{r}{6cm}
\centering
\includegraphics[width=6cm]{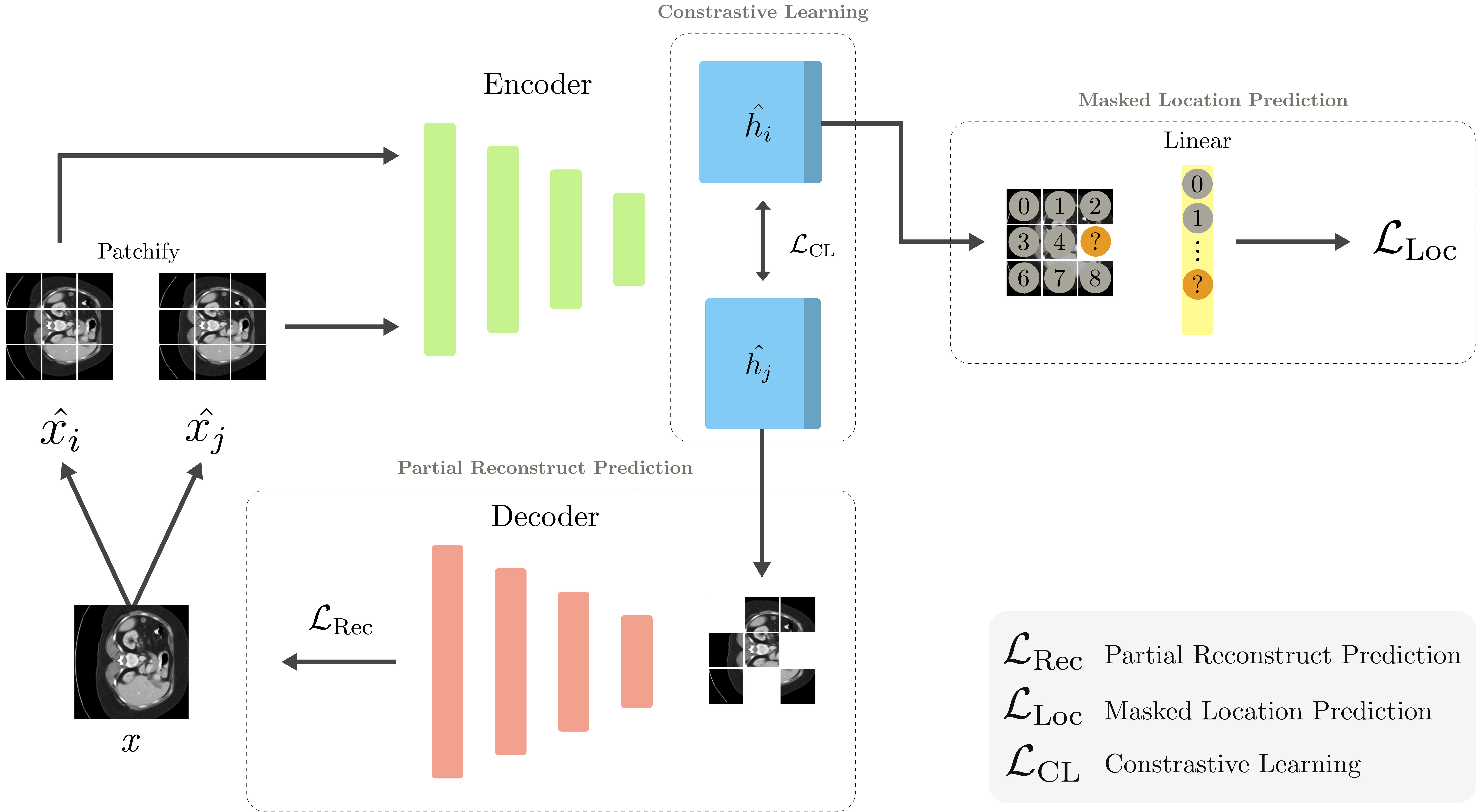}
\vspace{-0.4cm}
\caption{\color{violet}Our proposed SSL framework}
\label{fig:fig_ssl}
\vspace{-0.5cm}
\end{wrapfigure}

\noindent \textbf{Self-supervised learning (SSL)}
 can encode anatomical information of the human body in the image effectively. We propose three proxy tasks for learning comprehensive semantic representations within masked images without using labels. Our framework combines (1) Contrastive learning(e.g. SimCLR\cite{chen2020simple}), which encodes masked images to improve the ability to distinguish between different samples with hidden feature representations; (2) Masked Location Prediction, which predicts the location of the samples; and (3) Partial Reconstruct Prediction(\textit{e.g.} SimMIM\cite{xie2022simmim}), which learns the feature representation by reconstructing the masked patch area of each sub-volume.  These widely recognized self-supervised learning strategies are both straightforward and effective.

 When the input(demonstrated 2D image in Fig.\ref{fig:fig_ssl}) is divided into patches and then passed as input to the encoder twice, two sets of latent embeddings are obtained, and a contrastive learning is performed through constrastive loss\cite{oord2019representation} (Eq.\ref{eq:contrastive}). Then, masked location prediction is conducted to predict the number of randomly masked parts by dividing the $\hat{h_i}$ image into patches from 0 to 8 (Eq.\ref{eq:location}).
In addition, Partial Reconstruct Prediction is performed by masking the image of $\hat{h_j}$, reconstructing it through a decoder, and learning the difference from the original through $L_2$ loss (Eq.\ref{eq:reconstruct}).

\vspace{-0.2cm}
\begin{equation}
 \mathcal{L}_{\mathrm{CL}}=-\log \frac{\exp \left(\operatorname{sim}\left(x_i, x_j\right) / t\right)}{\sum_k^{2 N} \mathds{1}_{k \neq i, j} \exp \left(\operatorname{sim}\left(x_i, x_k\right) / t\right)}
 \label{eq:contrastive}
 \end{equation}

\vspace{-0.2cm}
 \begin{equation}
    \mathcal{L}_{\mathrm{Loc}} = - \frac{1}{R} \sum_{n=1}^{R} {v}_{n} log( \hat{v}_{n})
    \label{eq:location}
\end{equation} 

\vspace{-0.2cm}
 \begin{equation}
    \mathcal{L}_{\mathrm{Rec}} = \frac{1}{|\hat{R}|}\sum_{r \in \hat{R}} || y_{r} - \hat{y}_{r}||_2
    \label{eq:reconstruct}
 \end{equation}

Finally, We minimize total objective loss functions combining partial reconstruction prediction, masked location prediction and contrastive learning losses, as follows:
\vspace{-0.2cm}
 \begin{equation}
 \mathcal{L}_{\mathrm{total}}=\mathcal{L}_{\mathrm{Rec}} + \lambda_{1}  \mathcal{L}_{\mathrm{Loc}} + \lambda_{2} \mathcal{L}_{\mathrm{CL}}
 \label{eq:totalloss}
 \vspace{-0.2cm}
 \end{equation}

where $\lambda_1,\lambda_2$ are set to 0.1 and 0.01  respectively, as a result of empirical experiments.

\section{Experiments}
\noindent \textbf{Datasets.} The pre-training dataset consists of medical images sourced from partial accessible CT, MRI datasets encompassing 3,358, 6,970 subjects respectively. Notably, the pre-training step does not involve the utilization of annotations or labels on this dataset. The primary objective of this pre-training process is to enable the model to learn meaningful representations from the available image data, thus eliminating the need for manual annotation. 
The BTCV\cite{landman2015miccai} dataset comprises 3D abdominal multi-organ CT images from 30 cases, each associated with a specific form and featuring 13 multi class segmentation objectives.
The BraTS2021\cite{DBLP:journals/corr/abs-2107-02314} dataset includes 1,251 subjects of brain MRI images. Each image is annotated with three segmentation targets and encompasses four modalities T1, T1Gd, T2, and T2-FLAIR. 
The ISIC2018\cite{CASSIDY2022102305} dataset contains 2,594 dermoscopic images of skin lesions, each annotated by experts for segmentation purposes. 
The Cityscapes\cite{Cordts2016Cityscapes} dataset contains several urban street scenes for segmentation purposes and is used to test the generalization performance of our approach. It consists of 3475 semantically annotated train, val sets and 1525 test set.
Details about datasets are shown in the appendix.

\noindent \textbf{Implementation Details.}
Our architecture implemented in PyTorch and MONAI\footnote{\href{https://monai.io/}{https://monai.io/}}. For pre-training tasks, the reconstruction strategy is applied with a mask ratio of 0.4. Moving on to the fine-tuning phase, the AdamW optimizer\cite{loshchilov2019decoupled} is used with a weight decay $1e^{-3}$. The warm-up is set to $0.1$ of the total epochs, and the learning rate undergoes linear updates following the Cosine Annealing schedule\cite{loshchilov2017sgdr}. The loss function incorporates DICE loss\cite{Sudre_2017}, BCE loss, and MSE loss. Random flips, rotations, intensity scaling, and shifts were applied to augment the data. We set the number of diffusion steps as $1000$, and the sliding window overlap rate is $0.8$ until the final prediction. Preprocessing details for each dataset are provided in the appendix.

\noindent \textbf{Evaluation Metrics} are important to quantify the performance of the segmentation model. Two commonly used metrics are the dice similarity coefficient\cite{ZOU2004178}(Dice) and the Hausdorff distance\cite{232073}(HD). The evaluation metric are as define. $Y$ and $\hat{Y}$ represent the actual and predicted values in input units, and $g'$ and $p'$ represent the actual and predicted values of points on the surface.
\begin{equation}
\textrm{Dice} = \frac{2\sum_{i=1}^{I} Y_{i}\hat{Y}_{i} }{\sum_{i=1}^{I}Y_{i}+ \sum_{i=1}^{I}\hat{Y}_{i}},
\label{eq:dice_score}
\end{equation}

\vspace{-0.4cm}
\begin{equation}
\textrm{HD} =\max \{{\max _{g' \in G} \min _{p' \in P} } \|g'- p'\|, 
\max _{p' \in P} \min_{g' \in G} \|p'-g'\| \}.
\label{eq:hd_score}
\end{equation}

\begin{wraptable}{l}{5.5cm}
\vspace{-0.5cm}
\centering
\scriptsize
\resizebox{\linewidth}{!}{%
\vspace{-0.3cm}
\begin{tabular}{cc}
\multirow{2}{*}{Label smoothing} & Average Accuracy \\ \cline{2-2}
                                    & IoU $\uparrow$  \\ \hline
 \ \ LS & $83.72 \pm 0.08$ \\                                    
 \ \ N-ULS~\cite{galdran2020non} & $83.91 \pm 0.04$  \\
 \ \ SVLS\cite{islam2021spatially} & $83.79 \pm 0.06$ \\
 \ \ Ours*($k$-NLS) & $\textbf{84.19} \pm 0.04$ \\ \hline
\end{tabular}
}
\vspace{-0.7mm}
\caption{\color{violet}Comparison with the other LS}
\label{table:ls_ablation_table}
\vspace{-0.6cm}
\end{wraptable}
\noindent \textbf{Exploring the performance of Label Smoothing}
We concentrate on the performance of k-neighbor label smoothing and explore its applicability to general datasets with structural properties. We explore its applicability to a cityscapes\cite{Cordts2016Cityscapes} dataset with structural properties by utilizing only our baseline model and label smoothing. Compared with basic label smoothing(uniform), Non-Uniform Label Smoothing(NULS), and especially Spatially Varying Label Smoothing(SVLS), which applies label smoothing to neighboring pixels using weight matrix in the form of Gaussian kernel, we can see that our performance is superior. Previous methods compensate for the label’s uncertainty in image segmentation, but our methods further estimate the positional relationship between classes to improve prediction performance with a label smoothing method, emphasizing that this can be easily applied to other tasks.

\begin{wraptable}{l}{6cm}
\centering
\scriptsize
\vspace{-0.3cm}
\begin{tabular}{lccc}
\multirow{2}{*}{Loss Function}  &  \multicolumn{3}{c}{Average Accuracy}\\
\cmidrule{2-4}
&  Dice $\uparrow$ & HD $\downarrow$ \\
\midrule
 \ \ \ \ \ Scratch & $81.12$ & $5.11$   \\
\midrule
 \ \ $\mathcal{L_\text{CL}}$ & $81.21$ & $5.10$ \\
 \ \ $\mathcal{L_\text{Loc}}$ & $81.23$ & $5.10$ \\
 \ \ $\mathcal{L_\text{Rec}}$ & $81.56$ & $5.06$ \\
 \ \ $\mathcal{L_\text{Rec}} + \mathcal{L_\text{CL}}$ & $81.87$ & $4.95$ \\
 \ \ $\mathcal{L_\text{Rec}} + \mathcal{L_\text{Loc}}$ & $81.61$ & $4.98$ \\
 \ \ $\mathcal{L_\text{Rec}} + \mathcal{L_\text{CL}+} \mathcal{L_\text{Loc}}$ & $\mathbf{82.19}$ & $\mathbf{4.85}$ \\
\bottomrule
\end{tabular}

\vspace{0.2cm}
\caption{\color{violet}Ablation study of the pre-training objective function}
\label{table:sslloss_ablation_table}
\vspace{-0.3cm}
\end{wraptable}

\noindent \textbf{Efficiency of Self Supervised Objectives.} We conduct comprehensive ablation experiments on the BTCV dataset to evaluate the efficiency of self-supervised learning. In these experiments, we employed specific settings for calculating the loss, and the obtained results are presented in the Table.\ref{table:sslloss_ablation_table}. Notably, the $L_\text{Rec}$ is learned based on the pixel representation of input images,  $L_\text{Loc}$ (Masked Location Prediction) is learned by recognizing the location of the masked region, and $L_\text{CL}$ (Contrastive Learning) is focused on contrastive learning at two augmented sample level. The $L_\text{Rec}$ lies in its important role in understanding meaningful representation learning from medical images, as shown in experimental results. By employing these three loss functions, our self-supervised learning approach aims to capture intricate details at both pixel and region levels, enhancing the model's ability to extract meaningful features from the the inputs.

\vspace{-0.4cm}
\begin{wraptable}{l}{5.5cm}
\centering
\scriptsize
\resizebox{\linewidth}{!}{%
\begin{tabular}{lccc}
\multirow{2}{*}{Architecture}  &  \multicolumn{3}{c}{Average Accuracy}\\
\cmidrule{2-4}
&  Dice $\uparrow$ & HD $\downarrow$ \\
\midrule
 \ \ \ \ \ Scratch & $81.12$ & $5.11$   \\
\midrule
SSL  \\
 \ \ Encoder$\mathcal{_\text{Frozen}}$ & $83.17$ & $4.55$ \\
 \ \ Encoder$\mathcal{_\text{Trainable}}$ & $84.67$ & $4.11$ \\
RBA & $82.60$ & $4.74$   \\
\midrule
Ours* & $\textbf{86.72}$ & $3.48$ \\
\bottomrule
\end{tabular}
}
\vspace{-0.2cm}
\caption{\color{violet}Comparing morphology-driven learning strategies}
\label{table:architecture_ablation}
\vspace{-0.4cm}
\end{wraptable}

\noindent \textbf{Selecting the optimal architecture}
Remind the our approaches, in the case of self-supervised learning (SSL), feature representations pre-trained from the three proxy tasks are transferred to the conditional encoder to assist in understanding the original image. We experiment with the effect of freezing or leaving all weights trainable during benchmark fine-tuning. Additionally, our framework explores the ablation study on reverse boundary attention, which is integrated with the general diffusion segmentation process. We comprehensively verify the effectiveness of morphology-driven learning within the architecture to prove its hypothesis. As shown in the Table.\ref{table:architecture_ablation}, the single module experiments(presented in the second section) show higher performance than the scratch model, but learning by leaving the conditional encoder trainable shows a large margin in the BTCV dataset. This indicates that feature representation was achieved by aligning the learned features well with the downstream task. Our model, which comprehensively combines morphology-driven learning techniques, shows remarkable improvement in results, and our final architecture is shown in Fig.~\ref{fig:fig2}.

\vspace{-0.8cm}
\begin{table}[b]
\scriptsize
\resizebox{\textwidth}{!}{%
\begin{tabular}{l|clccccccccc|c}
\hline
Method                                                      & \multicolumn{1}{l}{Spleen} & Kidney & \multicolumn{1}{l}{Gall} & \multicolumn{1}{l}{Esophagus} & \multicolumn{1}{l}{Liver} & \multicolumn{1}{l}{Stomach} & \multicolumn{1}{l}{Aorta} & \multicolumn{1}{l}{IVC} & \multicolumn{1}{l}{Veins} & \multicolumn{1}{l}{Pancreas} & \multicolumn{1}{l|}{AG} & \multicolumn{1}{l}{Avg.} \\ \hline
TransUNet~\cite{chen2021transunet}   & 0.952& 0.928& 0.662& 0.757& 0.969& 0.889& 0.920& 0.833& 0.791& 0.775& 0.637& 0.828\\
nnUNet~\cite{isensee2018nnunet}       & 0.947& 0.920& 0.794& 0.812& 0.955& 0.905& 0.908& 0.850& 0.812& 0.829& 0.764& 0.863\\
UNETR~\cite{hatamizadeh2021unetr}     & 0.952& 0.928 & 0.805& 0.824& 0.963& \textbf{0.925}& 0.928& 0.857& 0.828& 0.832& 0.781& 0.874\\
Swin UNETR~\cite{hatamizadeh2022swin} & 0.956& 0.937& 0.828& 0.827& 0.971& 0.921& 0.928& 0.863& 0.849& 0.858& 0.810& 0.886\\ \hline
EnsemDiff~\cite{wolleb2021diffusion}  & 0.905& 0.911& 0.732& 0.723& 0.947& 0.838& 0.915& 0.838& 0.704& 0.715& 0.637& 0.805\\
SegDiff~\cite{amit2022segdiff}        & 0.894& 0.881& 0.703& 0.654& 0.852& 0.702& 0.874& 0.819& 0.715& 0.724& 0.694& 0.774\\
MedsegDiff~\cite{wu2023medsegdiff}    & 0.969& 0.930& 0.822& 0.817& 0.970& 0.919& 0.912& 0.859& 0.831& 0.813& 0.791& 0.875\\
Diff-UNet~\cite{xing2023diffunet}      & \textbf{0.973}& \textbf{0.942}  & 0.812& 0.815& \textbf{0.973}& 0.924& 0.907& 0.868& 0.825& 0.788& 0.779& 0.873\\ \hline
Ours*   & 0.972& \textbf{0.942}  & \textbf{0.903}& \textbf{0.847}& 0.972& 0.924& \textbf{0.945}& \textbf{0.874}& \textbf{0.867}& \textbf{0.880}& \textbf{0.842}& \textbf{0.906}\\ \hline
\end{tabular}
}
\vspace{-0.2cm}
\caption{\color{violet}Quantitative results for multi-organ segmentation. Note: Gall: gall bladder, IVC: inferior vena cava, AG: left and right adrenal glands.}
\label{tab:btcv}
\end{table}
\vspace{-0.5mm}
\begin{figure*}
\centering
\includegraphics[width=12cm,height=6cm]{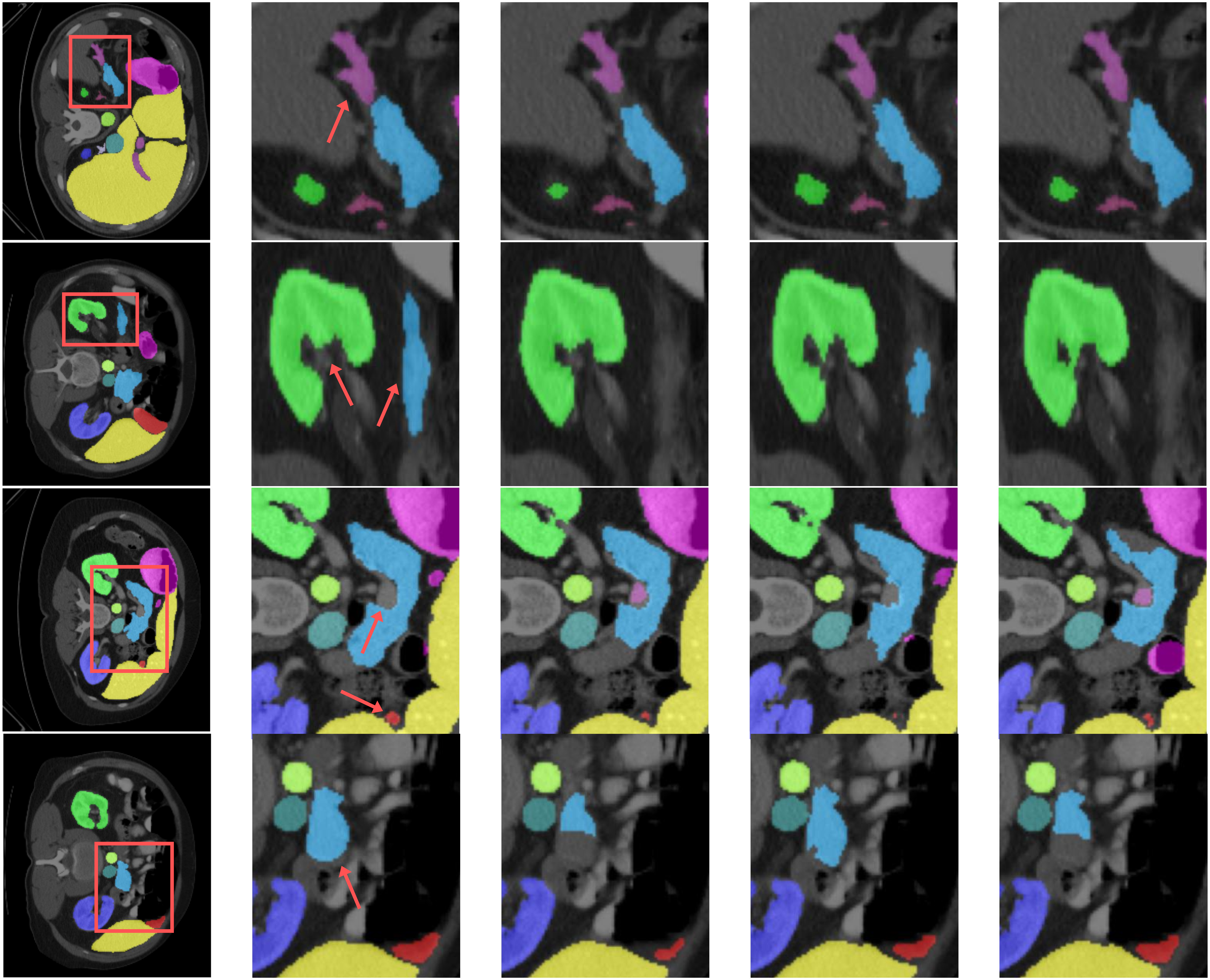}
\vspace{1.5mm}
\caption{\color{violet}Qualitative results of the proposed model. The region of interest was highlighted with arrows. (From left: GroundTruth, DTS(our), UNet, SwinUNETR, Diff-UNet)}
\label{fig:fig5}
\vspace{-1.7mm}
\end{figure*}

\section{Comparative Results}
As shown in Table.\ref{tab:btcv}, we compare our model with the BTCV benchmark dataset. Compared with other models, the proposed DTS achieves the best performance and presents a higher dice result of 0.906. It can be seen that previous diffusion segmentation models show comparable performance to conventional segmentation models in relatively large organs(\textit{e.g.} liver, stomach), but poor performance in small organs(\textit{e.g.} esophagus, aorta). DTS surpasses the closest competing methods by an average of 2\% across all classes, with an even more significant improvement of 7\% specifically for gall bladder. We believe that our approach and the application of the high performance transformer architecture will lead to improved accuracy. Comprehensive qualitative results of our model, which demonstrate good segmentation performance for small organs, can be found in Fig.\ref{fig:fig5}, highlighting our model's ability to capture details and achieve accurate boundary representations.

\begin{table*}[b]
\scriptsize
\centering
\resizebox{\textwidth}{!}{%
\begin{tabular}{c|cccccccc|cc}
& \multicolumn{8}{c|}{BraTs}                                                                                                            & \multicolumn{2}{c}{ISIC}        \\
& \multicolumn{2}{c}{WT}          & \multicolumn{2}{c}{TC}          & \multicolumn{2}{c}{ET}          & \multicolumn{2}{c|}{Average}    & \multicolumn{2}{c}{Average}     \\ \cline{2-11} 
\multirow{-3}{*}{\textbf{Method}} & Dice↑& HD↓& Dice↑& HD↓& Dice↑& HD↓& Dice↑& HD↓& Dice↑& HD↓\\ \hline
TransUNet~\cite{chen2021transunet}& 78.95& 5.87& 81.60& 5.05& 76.15& 5.91& 78.90& 5.87& 85.40& 3.88\\
UNETR~\cite{hatamizadeh2021unetr}& 89.92& 2.49& 84.79& 4.07& 79.51& 5.77& 84.74& 4.08& 87.57& 3.21\\
SwinUNETR~\cite{hatamizadeh2022swin}& \textbf{90.04}& \textbf{2.41}& 85.19& 3.94& 80.01& 5.69& 85.09& 3.97& 89.68& 2.57\\ \hline
SegDiff~\cite{amit2022segdiff}& 80.51& 5.23& 82.32& 4.83& 73.24& 6.84& 78.69& 5.87& 87.30& 3.32\\
MedsegDiff~\cite{wu2023medsegdiff}& 89.49& 2.71& 85.12& 3.96& 79.12& 5.81& 84.57& 4.13& 89.89& 2.57\\
Diff-UNet~\cite{xing2023diffunet}& 88.23& 2.94& 86.94& 3.40& 79.87& 5.79& 85.01& 4.01& 88.64& 2.94\\ \hline
Ours* & 89.63& 2.57& \textbf{88.02}& \textbf{3.07}& \textbf{81.11}& \textbf{5.12}& \textbf{86.25}& \textbf{3.62}& \textbf{91.12}& \textbf{2.18} \\ \hline

\end{tabular}}
    
\vspace{1.9mm}
\caption{\color{violet}Quantitative result on BraTS and ISIC dataset. Note: WT: whole Tumor, TC: tumor core, ET: enhancing tumor}
\label{tab:brats_isic}
\vspace{-0.5mm}
\end{table*}
The results presented in Table.\ref{tab:brats_isic} demonstrate that the two datasets showed optimal outcome with an average accuracy in terms of both Dice and HD score. In particular, within the ISIC dataset, solely K-neighbor label smoothing was omitted from the application. This decision was made due to the dataset has only a single label without structural position relationships between adjacent labels. Consequently, employing the K-neighbor label smoothing method in this specific scenario is unnecessary. Overall, SwinUNETR~\cite{hatamizadeh2022swin} has a competitive performance in the benchmark results. Although it employs an architecture similar to DTS, which facilitates the learning of multi-scale contextual information through a hierarchical encoder with a self-attention module, thereby effectively modeling long-range dependencies, it does not achieve the same level of robustness. This is because diffusion models excel at handling noise and artifacts in input data, particularly in medical images.

\section{Future work and Conclusion}
Our study focuses on the advantages of morphology-driven learning for segmentation tasks, where our approach demonstrates substantial improvements. Building on these promising results, we aim to broaden the scope of our framework by applying it to other critical imaging tasks, such as classification and detection, to evaluate its effectiveness across various domains and imaging scenarios. Moreover, we compare the performance of conventional segmentation models with diffusion-based models and plan to extend this analysis to include a detailed evaluation of multimodal large language models (MLLMs). This allows us to explore the potential advantages and limitations of models in the context of segmentation tasks, providing a broader understanding of their effectiveness.
In conclusion, we present a novel approach to medical image segmentation. DTS suggests the potential to replace existing CNN-based down-sampling by using a Swin Transformer encoder. We believe that this model architecture enables accurate segmentation with small, detailed representations and improves performance by complementing the chronic problems of medical images with Morphological-based learning, such as k-neighbor label smoothing, reverse boundary attention and self-supervised learning. We hope that this inspires future tasks in situations with morphologically complex problems.



\section*{Acknowledgements}
This research was supported by the MSIT(Ministry of Science and ICT), Korea, under the ITRC(Information Technology Research Center) support program(IITP-2024-2020-0-01789), and the Artificial Intelligence Convergence Innovation Human Resources Development (IITP-2024-RS-2023-00254592) supervised by the IITP(Institute for Information \& Communications Technology Planning \& Evaluation). Additionally, this work has been carried out with (extensive) use of the NEURON computing resource that is supported by the Korea Institute of Science and Technology Information (KISTI).

\bibliography{egbib}
\end{document}